\documentclass[lettersize,journal]{IEEEtran}
\usepackage{amsmath,amsfonts}
\usepackage{algorithmic}
\usepackage{algorithm}
\usepackage{array}
\usepackage[caption=false,font=normalsize,labelfont=sf,textfont=sf]{subfig}
\usepackage{textcomp}
\usepackage{stfloats}
\usepackage{url}
\usepackage{verbatim}
\usepackage{graphicx}
\usepackage{cite}
\begin{document}
\title{MAUNet-Light: A Concise MAUNet Architecture for Bias Correction and Downscaling of Precipitation Estimates}

\author{Sumanta Chandra~Mishra Sharma,  Adway ~Mitra ,~\IEEEmembership{Member,~IEEE,} Auroop Ratan Ganguly,~\IEEEmembership{Senior Member,~IEEE}

\thanks{S.C. Mishra Sharma is with the the Department of Artificial Intelligence, Indian Institute of Technology Kharagpur, India. e-mail: (sumantamishra22@gmail.com)}
\thanks{A. Mitra is with the Department of Artificial Intelligence, Indian Institute of Technology Kharagpur, India. e-mail: (adway@cai.iitkgp.ac.in)} 
\thanks{A R Ganguly is with Sustainability and Data Sciences Laboratory (SDS Lab), Civil and Environmental Engineering, Northeastern University, Boston, MA. USA and AI for Climate and Sustainability (AI4CaS), The Institute for Experiential AI (IEAI), Northeastern University, Boston, MA and Portland, ME, USA, (a.ganguly@northeastern.edu)}}
\maketitle 
   
\begin{abstract}
Satellite-derived data products and climate model simulations of geophysical variables like precipitation, often exhibit systematic biases compared to in-situ measurements. Moreover, their resolution is also relatively coarse-grained, upto $0.25^{\circ}$ for most models. Bias correction and spatial downscaling are fundamental components to develop operational weather forecast systems, as they seek to improve the consistency between coarse-resolution climate model simulations or satellite-based estimates and ground-based observations. In recent years, deep learning–based models have been increasingly replaced traditional statistical methods to generate high-resolution, bias free projections of climate variables. For example, Max-Average U-Net (MAUNet) architecture has been demonstrated for its ability to downscale precipitation estimates. The versatility and adaptability of these neural models make them highly effective across a range of applications, though this often come at the cost of high computational and memory requirements. The aim of this research is to develop light-weight neural network architectures for both bias correction and downscaling of precipitation, for which the teacher-student based learning paradigm is explored.  This research demonstrates the adaptability of MAUNet to the task of bias correction, and further introduces a compact, lightweight neural network architecture termed MAUNet-Light. The proposed MAUNet-Light model is developed by transferring knowledge from the trained MAUNet, and it is designed to perform both downscaling and bias correction with reduced computational requirements without any significant loss in accuracy compared to state-of-the-art. The number of parameters required for MAUNet-Light is approximately 60\% of the number of parameters needed for MAUNet. Experimental results show that MAUNet-Light effectively corrects systematic biases in satellite-based TRMM precipitation estimates by calibrating them against ground-based gridded precipitation data from the India Meteorological Department (IMD). In addition, the model is evaluated for precipitation downscaling of Indian Summer Monsoon Rainfall and is found to perform effectively for this task as well. This work demonstrates how light-weight neural models can be used effectively in the climate domain through appropriate pre-training.
\end{abstract}

\vskip 0.3in
\section{Introduction}
\IEEEPARstart{A}{ccurate} precipitation estimation is crucial for a wide range of hydrological and meteorological applications, including flood forecasting, drought monitoring, and climate change studies\cite{ref1_Wang2021}. However, precipitation estimates from different climate forecast models often struggle to capture local climatic conditions due to their coarse resolution. On the other hand, satellite-based precipitation estimates(SPEs), such as those provided by the Tropical Rainfall Measuring Mission (TRMM) and the Integrated Multi-satellite Retrievals for GPM (IMERG), offer extensive spatial and temporal coverage, making them invaluable in regions with sparse ground-based observations. Nonetheless, these satellite estimates often suffer from biases caused by factors such as retrieval algorithms, sensor limitations, and atmospheric conditions\cite{ref2_Maggioni2022}\cite{ref3_Le2023}. Therefore, it is essential to downscale coarse resolution climate model forecasts and correct the biases present in model- or satellite-based precipitation estimates to generate more accurate, high-resolution precipitation data for informed local decision-making.

To produce high-resolution climate maps and to correct the bias of SPEs different statistical\cite{ref4_Wilby2021}\cite{ref5_Passow2020}\cite{ref6_Guo2019} and deep learning based techniques\cite{ref7_Wang2023}\cite{ref8_Mishra_Sharma2022}\cite{ref9_Chen2022}\cite{ref10_Dai}\cite{ref11_Mishra_Sharma2024_CI} have been employed. Statistical techniques used for this task typically fall into two broad categories, namely mean-based approaches\cite{ref12_Luo2020} and distribution-based approaches\cite{ref13_Katiraie2020}\cite{ref14_Irwandi}. These methods often apply scaling and transformation operations to generate high-resolution, bias-corrected results. However, these statistical methods often fail to capture the complex, nonlinear relationships between input features and the target variable. In contrast, deep learning-based models \cite{ref15_Sun2021}\cite{ref16_Li2025} utilize advanced neural network architectures to learn intricate mappings between model or satellite estimates and ground-based observations. This enables more accurate and robust solutions for downscaling and bias correction\cite{ref17_Wang2022}\cite{ref18_Mishra_Sharma2024_AR}.

Deep neural architectures developed for specific tasks can often be adapted for related tasks through training on new datasets and leveraging transfer learning. The versatility and adaptability of these architectures make them highly effective across a range of applications. Also, the introduction of model compression in deep learning has initiated the development of lightweight models for resource-constrained environments\cite{ref19_Lyu2024}\cite{ref20_Bhalgaonkar2024}. The lightweight models developed by applying different model compression techniques\cite{ref21_Zhu2018}\cite{ref22_Menghani2023}\cite{ref23_Gou2021}\cite{ref24_Zhang18} are computationally efficient as well as memory efficient. These models aim to produce relevant results without compromising performance too much\cite{ref25_Dantas2024}.

Model compression is relatively rare in deep learning based downscaling and bias correction of climate variables. The available literature shows that model compression and the development of lightweight models for these applications remain largely unexplored, presenting a promising direction for research. Therefore, this research aims to develop compressed models for downscaling and bias correction with the help of student-teacher based training with knowledge refinement.

In our previous research, we introduced the MAUNet model based on Max-Average Pooling for downscaling gridded precipitation data to higher resolutions\cite{ref26_Mishra_Sharma2024_MAUNet}. The term MAUNet used in this work represents the MAUNet(D) architecture. In this work, we extend MAUNet to address the task of bias correction for satellite-based precipitation estimates. Additionally, we present a compact, light-weight variant of MAUNet, termed MAUNet-Light, designed for bias correction and downscaling of precipitation estimates. The proposed MAUNet-Light is trained using a Teacher-Student learning framework, where the model learns from the predictions of a pre-trained teacher model and is subsequently calibrated using ground-truth data. 
This study focuses on the bias correction of satellite-based TRMM precipitation estimates by calibrating them against ground-based IMD precipitation data. It also employs the proposed architectures to downscale IMD gridded precipitation estimates.

The rest of this article is organized as follows: Section II presents the datasets used in this research. Section III describes the methodologies, including the model architecture and training procedures of the proposed models. In Section IV, we evaluate the performance of the models and provide a comparative analysis of the results. Finally, Section V concludes the study with a summary of the key findings.

\section{Dataset}
This research utilizes the TRMM\_3B42\_Daily rainfall dataset \cite{ref27_Huffman2016} and IMD gridded rainfall dataset \cite{ref28_Pai2014} to carry out the bias correction task. The TRMM\_3B42\_Daily dataset is produced by NASA GES DISC from the research-quality 3-hourly TRMM Multi-Satellite Precipitation Analysis (TMPA\_3B42). Similarly, the IMD daily gridded rainfall dataset is prepared by the India Meteorological Department from rain gauge-based ground observations. These datasets provide gridded rainfall values with a spatial resolution of 0.25° × 0.25° and temporal resolution of 1 day. This study focuses on the bias correction of TRMM gridded precipitation estimates for the ISMR period, using the IMD gridded rainfall dataset as the ground observation. The research exclusively focuses on precipitation data over the mainland of India, spanning the latitude range of 6.75°N to 38.5°N and the longitude range of 66.5°E to 98.25°E. The shape of the data samples used in this work is $128 \times 128$.
For the bias correction task, we gathered data samples from 1998 to 2019, utilizing data from 1998 to 2014 for training the deep learning models and the remaining data for testing. For the quantile-based mapping approaches, the calibration period is set from 2010 to 2014, and the projection period extends from 2015 to 2019. 

For the downscaling task, we use the IMD gridded rainfall dataset at two spatial resolutions: 1° × 1° (low-resolution)\cite{ref29_Rajeevan} and 0.25° × 0.25° (high-resolution)\cite{ref28_Pai2014}. In this work, we adopt the same data samples as used in \cite{ref26_Mishra_Sharma2024_MAUNet}, ensuring consistency and comparability with prior studies. Each low-resolution sample has a shape of $35\times35$, while the corresponding high-resolution (observed) sample is $140\times140$. Data from 1990–2012 is used for training, and 2013–2019 for testing.

Both the bias correction and downscaling tasks performed here concentrate on the precipitation data of mainland of India. Therefore, while preparing the training samples, we have considered a `0' value for the outside grid locations for both the TRMM and IMD data samples.

\section{Methodology}
\subsection{MAUNet}
\begin{figure}[h]
\vskip 0.1in
\begin{center}
\centerline{\includegraphics[height=4.5in, width=3.2in]{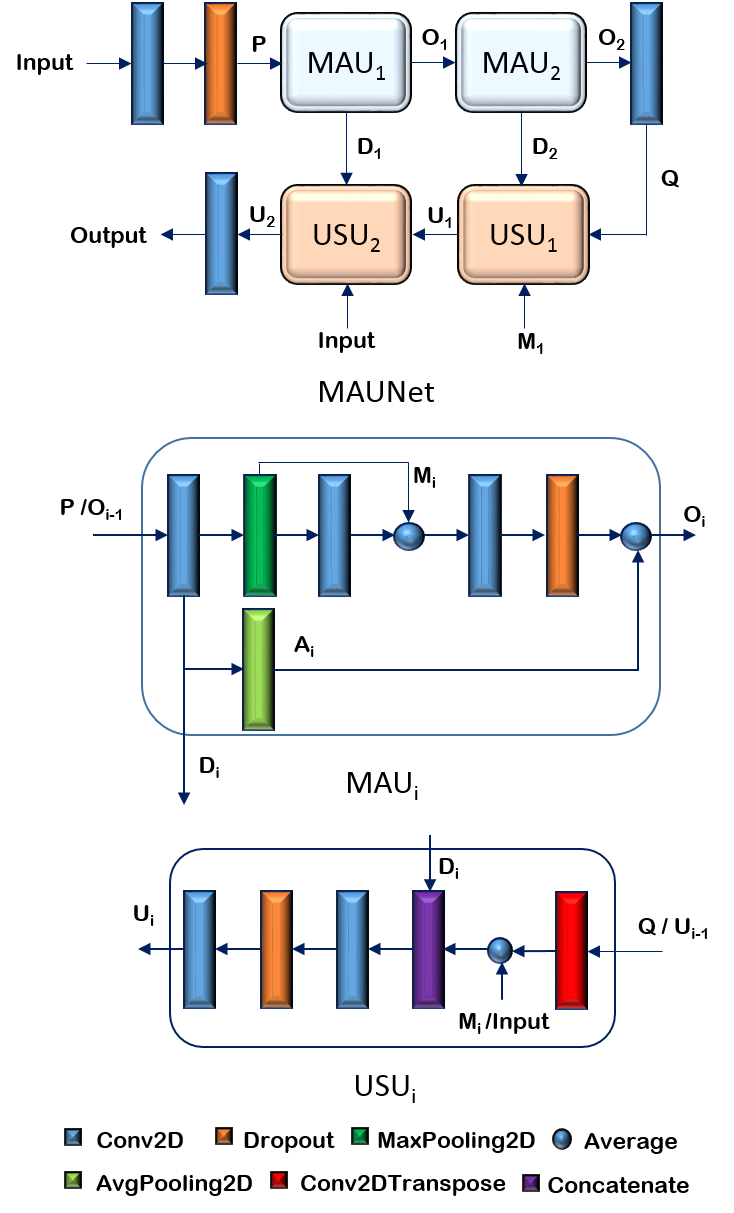}}
\caption{Schematic diagram of the MAUNet architecture.}
\label{MAUNet}
\end{center}
\vskip -0.2in
\end{figure}

MAUNet is a Max-Average UNet architecture (Fig. \ref{MAUNet}) that was previously used to produce high-resolution climate projections from coarse resolution data\cite{ref26_Mishra_Sharma2024_MAUNet}. It is a pre-upsampled model that refines the linearly interpolated (upsampled) input to produce a clean high-resolution output. The task of refining the upsampled input with MAUNet closely resembles the bias correction approach. Hence, in this work, we explore the usability of MAUNet in bias correction.

MAUNet contains two Max-Average Units (MAU) and two Upsampler Units (USU) along with other layers. The MAU extracts important features from the data, while the USU constructs refined samples. The combined use of max-pooling and average-pooling in MAUNet architecture helps in capturing the important pixel values while maintaining smoothness of the results. Additionally, the model includes skip connections and averaging units that controls gradient flow in the network.

In this network, the first convolutional layer produces 64 feature maps, while the convolutional layers of $MAU_{1}$ and $MAU_{2}$ produce 32 and 16 feature maps, respectively. Similarly, the Conv2D layers of $USU_{1}$ have 32 channels, while the Conv2D layers of $USU_{2}$ have 16 channels each. The intermediate layer present between $MAU_{2}$ and $USU_{1}$ has 16 filters and produces the output `Q' which is treated as input for $USU_{1}$. The final convolutional layer of this model generates a single feature map corresponding to the downscaled or bias-corrected result for the specified input.

Let the input-to-output mapping of $MAU_{i}$ and $USU_{i}$ be functionally represented as, $y=f_{i}(x)$ and $y=g_{i}(x)$ respectively for input $x$ and output $y$ of these units. Similarly, if $Conv(x)$ represents the convolutional operation on $x$ and $Drop(x,\alpha)$ indicates the dropout operation on $x$ with dropout percentage $\alpha$, then the input-to-output mapping of the MAUNet model can be presented with the help of the following equations. 

\begin{equation}
P=Drop(Conv(X),\alpha = 30\%)
\end{equation}
\begin{equation}
Q=Conv(f_{2}(f_{1}(P)))
\end{equation}
\begin{equation}
U_{i}=g_{i}(Z_{i})
\end{equation}
where, $Z_{i} = \Bigg\{  \begin{tabular}{ccc}
	  $(Q,D_{2},M_{1})$ for $i=1$ \\
	  $(U_{1},D_{1},X)$ for $i=2$ \\
	  \end{tabular}$

and,
\begin{equation}
Y=Conv(U_{2})
\end{equation}
In these equations, $X$ and $Y$ represent the input and output of the MAUNet model, while $P$, $Q$ and $U_{i}$ denote the output of the intermediate layers of the model. Here, $Z_{i}$ represents the input received by $USU_{i}$.

\subsection{MAUNet-Light}
\begin{figure}[ht]
\begin{center}
\centering
\centerline{\includegraphics[height=2.8in, width=3.3in]{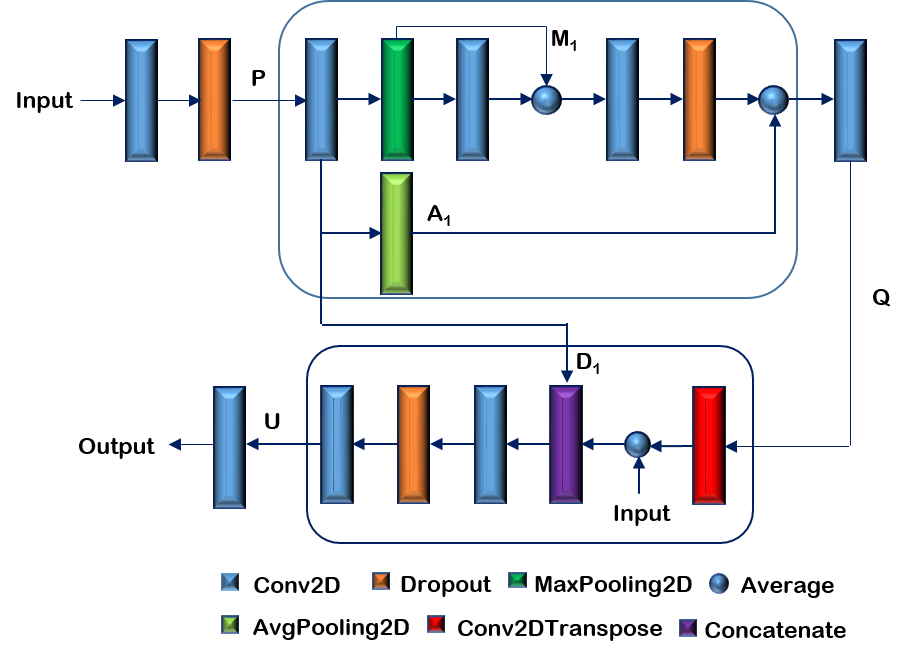}}
\caption{Schematic diagram of the MAUNet-Light architecture.}
\label{MAUNet_Light}
\end{center}
\vskip -0.2in
\end{figure}

The MAUNet-Light architecture presented in Fig. \ref{MAUNet_Light} is a compressed version of MAUNet. This model is prepared by considering a single Max-Average Unit and a single Upsampler Unit along with other relevant layers. Like MAUNet the first convolutional layer of MAUNet-Light acts as a feature extraction layer and produces 64 feature maps from the single channel input. These feature maps are passed through the dropout layer with $\alpha=30\%$ before entering the MAU unit. The convolutional layers of the MAU unit of this model have 32 filters, while the convolutional layers of the USU unit have 16 filters each. The dropout layer present in the MAU unit has $\alpha=20\%$, and the dropout layer of the USU unit has $\alpha=10\%$. In terms of functional notation, the operation of MAUNet-Light can be presented as,

\begin{equation}\label{eq:5}
P=Drop(Conv(X^*),\alpha = 30\%)
\end{equation}
\begin{equation}\label{eq:6}
Q=Conv(f_{1}(P))
\end{equation}
\begin{equation}\label{eq:7}
U=g_{1}(Q,D_{1},X^*)
\end{equation}
and,
\begin{equation}\label{eq:8}
Y^*=Conv(U)
\end{equation}
In these equations, $X^*$ and $Y^*$ represent the input and output of the MAUNet-Light
model, while the other notations retain the same meanings as defined earlier. In Equation~\ref{eq:7}, the function $g_{1}()$ takes three arguments: $Q,D_{1}$ and $X^*$. These serve as inputs to different layers of the upsampler unit. This equation highlights the presence of a long skip connection in the network, which enables the model to learn the residual more effectively.

In terms of complexity, MAUNet-Light seems to be more simpler than MAUNet. The comparison of these two models in terms of their complexity is presented in Table \ref{ch7:Table_MAUNet_Complexity_Comperision}. The table shows that the number of parameters required for MAUNet-Light is approximately $60\%$ of the number of parameters needed for MAUNet. Additionally, MAUNet-Light is more efficient than MAUNet in terms of memory requirements and floating point operations.

\vskip -0.1in
\begin{table}[h]
\caption{Comparison of MAUNet and MAUNet-Light}
\vskip -0.15in
\label{ch7:Table_MAUNet_Complexity_Comperision}

\begin{center}
\begin{tabular}{|p{1.8cm}|p{1.5cm}|p{2.2cm}|p{1.2cm}|}
\hline
\textbf{Model} & \textbf{Number of Parameters} & \textbf{Memory Space Occupied by Parameters} & \textbf{FLOPs} \\
\hline
MAUNet        & 85697 & 334.75KB & 1.3437G \\
MAUNet\_Light  & 52529 & 205.19KB & 1.1273G  \\

\hline
\end{tabular}
\end{center}
\vskip -0.3in
\end{table}

\subsection{Training the Models with Ground-truth Recalibration}
In this experiment, we have generated four trained models by training the MAUNet and MAUNet-Light Model with different input and target set. First, we train the MAUNet model by considering the respective input and target samples. During training this model, we have considered MSE as a loss function and ADAM as the optimizer. We used an early stopping criterion with patience = 20 for training the model. The trained MAUNet model is referred to as the teacher model.

Like the MAUNet model, we have also trained the MAUNet-Light model by considering the same input and target samples as used in MAUNet training. The resulting model from this training is named MAUNet\_Light\_GT, where `GT' indicates that it has been trained using the actual ground truth (HR IMD Data).

In the third case, we utilized the trained MAUNet model to generate output samples for the training duration. These predicted output samples are called MAUNet\_Prediction or Teacher\_Prediction. Subsequently, we trained a new MAUNet-Light model using TRMM (or LR IMD Data) as input and MAUNet\_Prediction as the target. This training approach can be considered as a Teacher-Student learning method, where the weights of the student model are updated based on the Teacher\_Prediction target. The student model prepared after this training is called MAUNet\_Light\_MP. Here MP indicates that the model is trained using the MAUNet\_Prediction as target.

\begin{figure}[h]
\begin{center}
\centerline{\includegraphics[height=5.4in, width=\columnwidth]{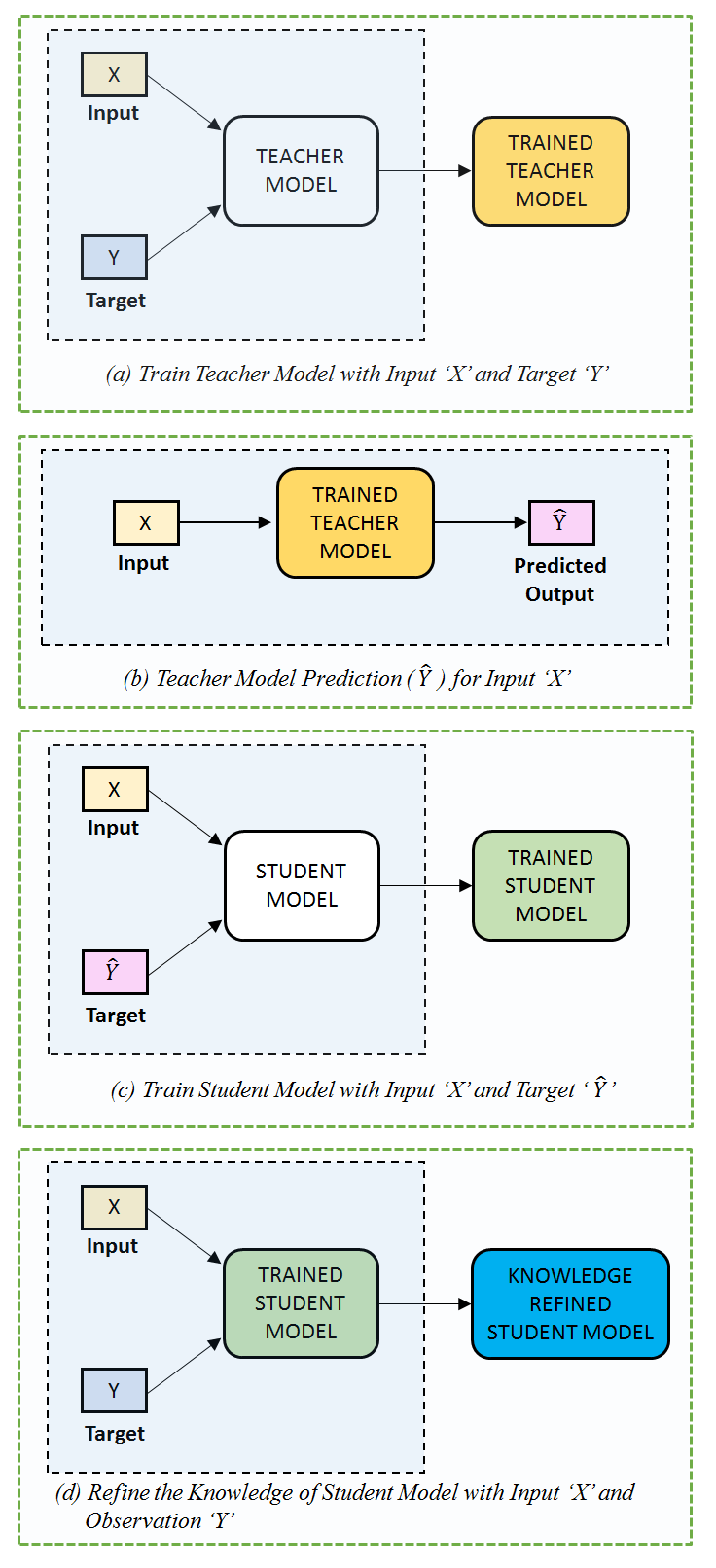}}
\caption{Overview of the MAUNet-Light training process.}
\label{ch7:MAUNet_Light_Training}
\end{center}
\vskip -0.3in
\end{figure}

Then in the fourth case, we have used a recalibration approach, where the trained weights of MAUNet\_Light\_MP model are refined by considering TRMM(or LR IMD Data) as the input and IMD observations as the target. This approach is called knowledge refinement, since we are refining the knowledge of the student model to improve its performance. The model prepared after this knowledge refinement is called MAUNet\_Light\_KR. The overall tarining approach of MAUNet\_Light\_KR is presented in Fig. \ref{ch7:MAUNet_Light_Training}.

In this experiment, the MAUNet-Light models are trained using MSE as the loss function and ADAM as the optimizer. Each of these training approaches uses an early stopping criterion with patience = 20. 

\section{Result and Discussion}

\subsection{Performance Evaluation for Downscaling}
In this subsection, we have analyzed the downscaling performance of the proposed lightweight models. We conducted a comparative analysis between our model and other state-of-the-art approaches. The performance evaluation is performed by considering the predicted and observed rainfall measures of all the valid grid locations in the test set. To assess the model's downscaling performance, we utilized four different performance metrics: RMSE, PSNR, MSSIM, and the correlation coefficient. The results for these metrics across various models are presented in Table \ref{Ch7_Table_Downscaling_Performance}.

\begin{table}[h]
\centering
\caption{Comparison of models in downscaling} \label{Ch7_Table_Downscaling_Performance}
\begin{tabular}{|l|l|l|l|l|}
\hline
\textbf{Model} & \textbf{RMSE} & \textbf{PSNR} & \textbf{MSSIM} & \textbf{Corr. Coef.} \\
\hline
Bilinear Interpolation & 13.6167 & 35.6124 & 0.8435 & 0.6230 \\
Bicubic Interpolation  & 14.6210 & 34.9943 & 0.8296 & 0.6137 \\
EDSR                   & 11.8247 & 36.8381 & 0.8646 & 0.6997 \\
SRDRN                  & 11.3761 & 37.1740 & 0.8779 & 0.7315 \\
MAUNet                 & 11.1947 & 37.3137 & 0.8752 & 0.7322 \\
MAUNet\_Light\_GT       & 11.5365 & 37.0524 & 0.8665 & 0.7107 \\
MAUNet\_Light\_MP       & 11.4515 & 37.1167 & 0.8717 & 0.7197 \\
MAUNet\_Light\_KR       & 11.4286 & 37.1341 & 0.8725 & 0.7199 \\
\hline
\end{tabular}
\end{table}

\begin{figure}[h]
\begin{center}
\centerline{\includegraphics[width=0.9\columnwidth]{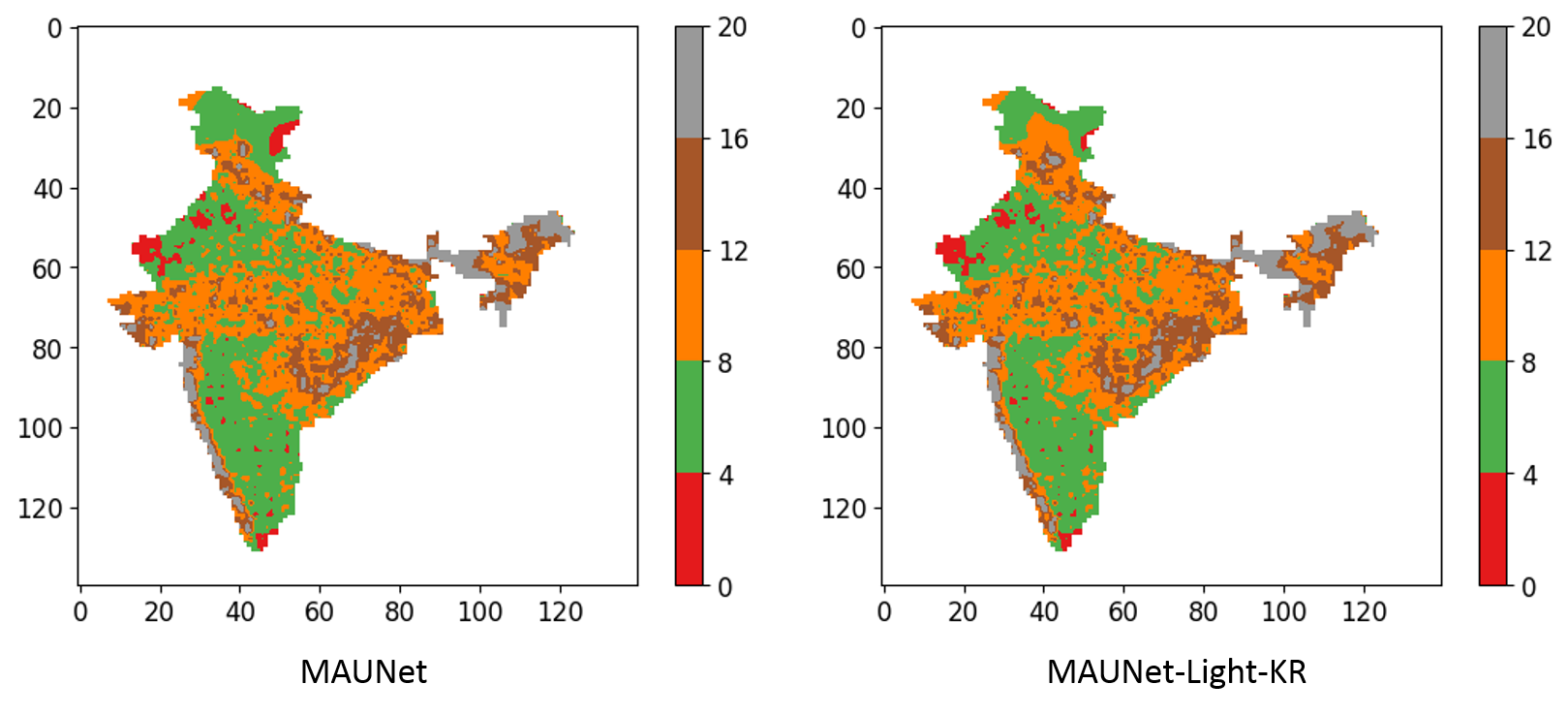}}
\caption{Gridwise RMSE plots for MAUNet and MAUNet\_Light\_KR in downscaling.}
\label{ch7:Gridded_RMSE_Plot_Downscaling}
\end{center}
\vskip -0.3in
\end{figure}

Our findings indicate that the lightweight model developed through knowledge refinement achieves performance close to that of the MAUNet model. To further strengthen our analysis, we calculated the grid-wise RMSE values for both the MAUNet and MAUNet\_Light\_KR models, which are illustrated in Fig. \ref{ch7:Gridded_RMSE_Plot_Downscaling}. This figure shows that the gridded RMSE values for the proposed lightweight model exhibit a similar spatial distribution pattern to that of the RMSE values obtained for the MAUNet model. For most grid points, the MAUNet\_Light\_KR produces an error rate that is comparable to that of the MAUNet.

\subsection{Performance Evaluation for Bias Correction}
\begin{figure*}[t!]
\begin{center}
\centerline{\includegraphics[width=1.8\columnwidth]{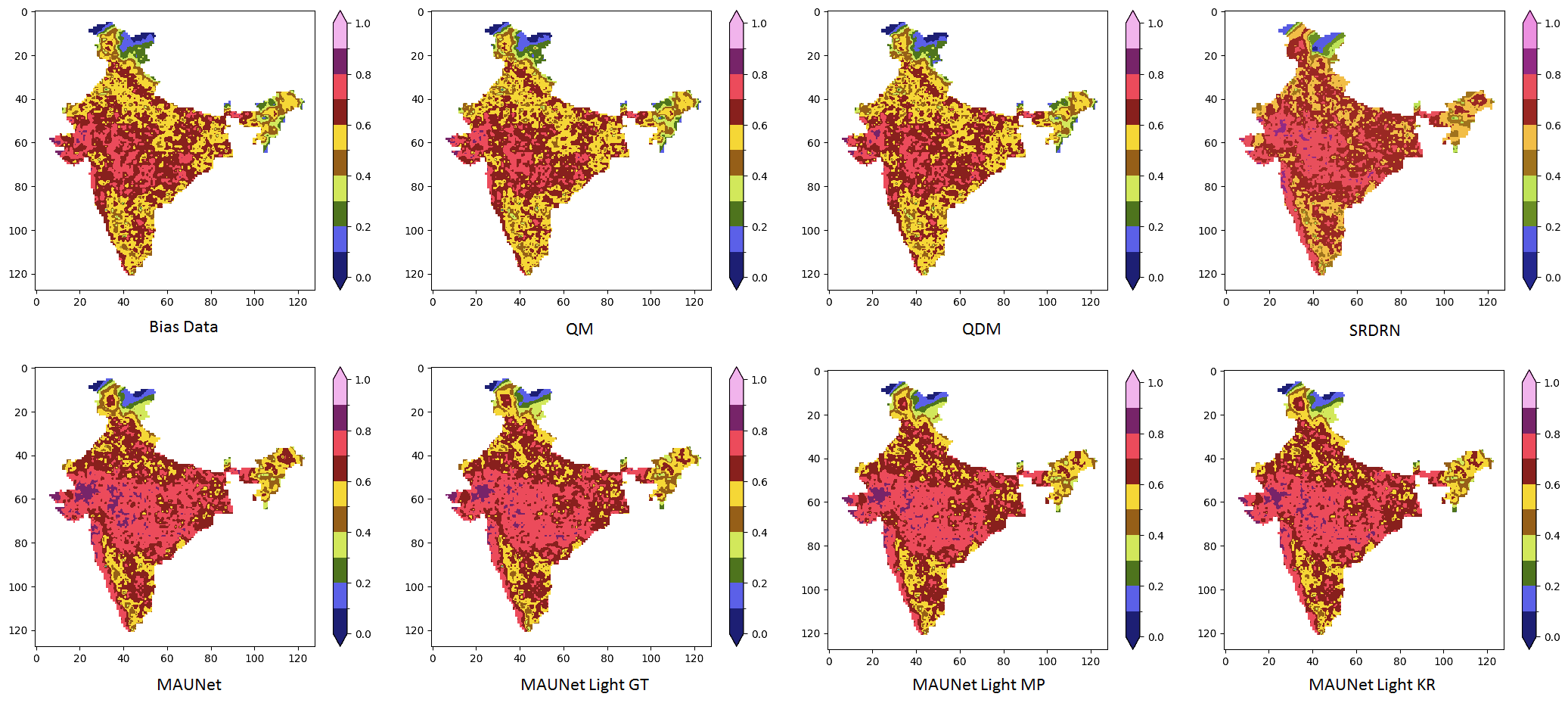}}
\vskip -0.1in
\caption{Gridwise correlation coefficient values illustrating the spatial variation in correlation between model predictions and observations.}
\label{ch7:Gridded_Correlation_Plot}
\end{center}
\vskip -0.3in
\end{figure*}

\begin{figure*}[b!]
\begin{center}
\centerline{\includegraphics[ width=1.8\columnwidth]{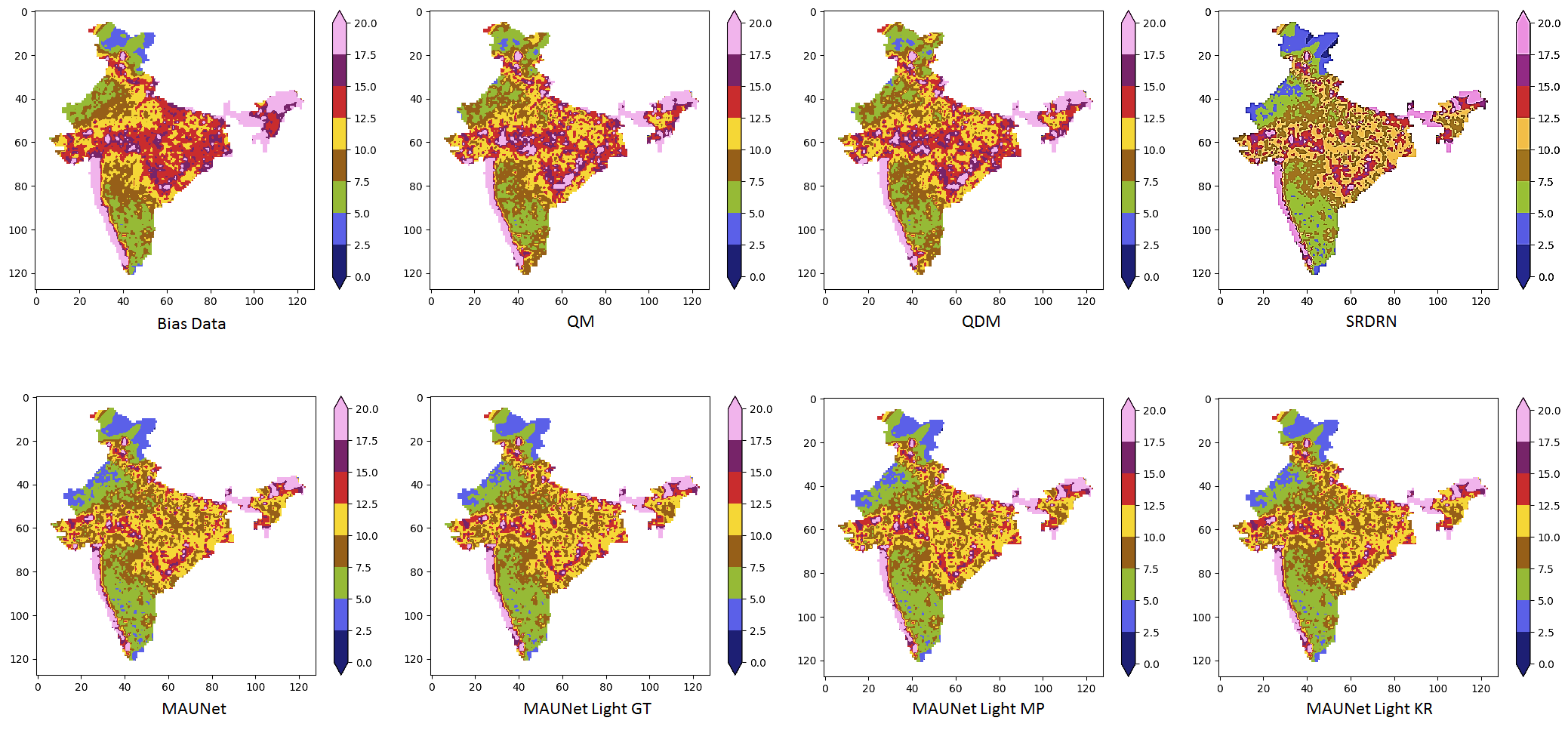}}
\vskip -0.1in
\caption{Gridwise RMSE values illustrating the spatial variation in model prediction errors.}
\label{ch7:Gridded_RMSE_Plot}
\end{center}
\vskip -0.3in
\end{figure*}

In this subsection, we assess the effectiveness of both trained deep learning models and calibrated statistical techniques for the task of bias correction by using two well-known performance metrics: the root mean square error (RMSE) and Pearson’s correlation coefficient (R). Here, the performance metrics are computed individually for each valid grid point through a comparison of predicted values with ground observations. 
The correlation coefficient values obtained at each grid location are shown in Fig. \ref{ch7:Gridded_Correlation_Plot}. and the RMSE values calculated at each grid location for different bias correction approaches are shown in Fig. \ref{ch7:Gridded_RMSE_Plot}. The spatial mean of these gridded correlation coefficients and the gridded RMSE values is presented in Table \ref{ch7:Table_Bias_Correction_Result}. 

\begin{table}[h!]
\caption{Mean values of Gridded Correlation Coefficient and Gridded RMSE }
\vskip -0.15in
\label{ch7:Table_Bias_Correction_Result}
\begin{center}
\begin{tabular}{|l|l|l|}
\hline
\textbf{Data/Model} & \textbf{Mean Corr. Coeff.} & \textbf{Mean RMSE} \\
\hline
Bias Data          & 0.5751 & 12.7848  \\
QM                 & 0.5655 & 12.6847  \\
QDM                & 0.5654 & 12.6121  \\
SRDRN              & 0.6230 & 10.7530  \\
MAUNet             & 0.6400 & 10.4645  \\
MAUNet\_Light\_GT  & 0.6337 & 10.5602  \\
MAUNet\_Light\_MP  & 0.6348 & 10.5626  \\
MAUNet\_Light\_KR  & 0.6378 & 10.5199  \\

\hline
\end{tabular}
\end{center}
\vskip -0.1in
\end{table}

From Fig. \ref{ch7:Gridded_Correlation_Plot} and Fig. \ref{ch7:Gridded_RMSE_Plot}, it can be observed that the deep learning based models improve correlation and reduce error across almost all grid points in India. Notably, there is a significant improvement in correlation and bias reduction in eastern, western, and central India. From the figures and table, it is evident that MAUNet outperforms all other models in bias correction. Additionally, the performance of MAUNet\_Light\_KR closely matches the MAUNet performance. It is also observed that the MAUNet\_Light\_KR model performs better than the statistical approaches and other deep learning methods in bias correction. The knowledge refinement approach improves the model performance compared to the direct training approach. Therefore, the knowledge refinement approach proves to be an effective method for training compressed deep learning models.

\begin{figure*}[t!]
\begin{center}
\centerline{\includegraphics[width=1.8\columnwidth]{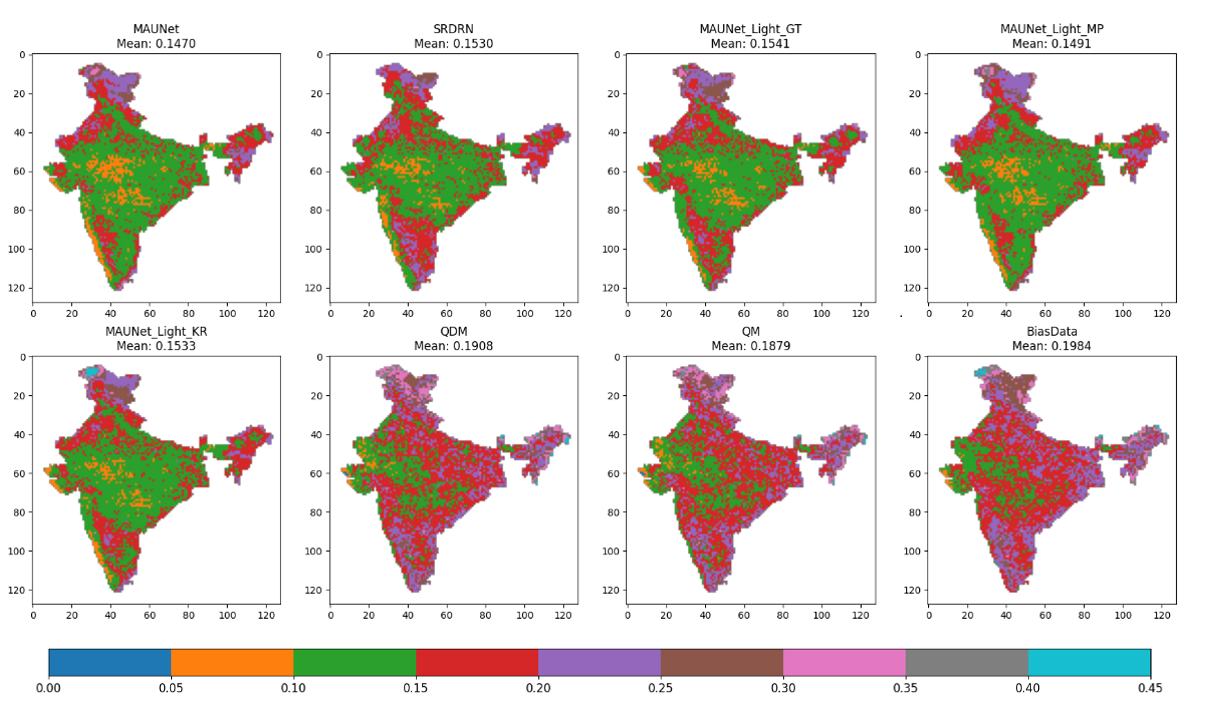}}
\vskip -0.1in
\caption{Gridwise KL divergence computed at each grid point, indicating distributional differences between model outputs and observations.}
\label{ch7:Gridded_KL_Divergence}
\end{center}
\vskip -0.3in
\end{figure*}

\begin{figure*}[h!]
\vskip -0.1in
\begin{center}
\centerline{\includegraphics[ width=1.6\columnwidth]{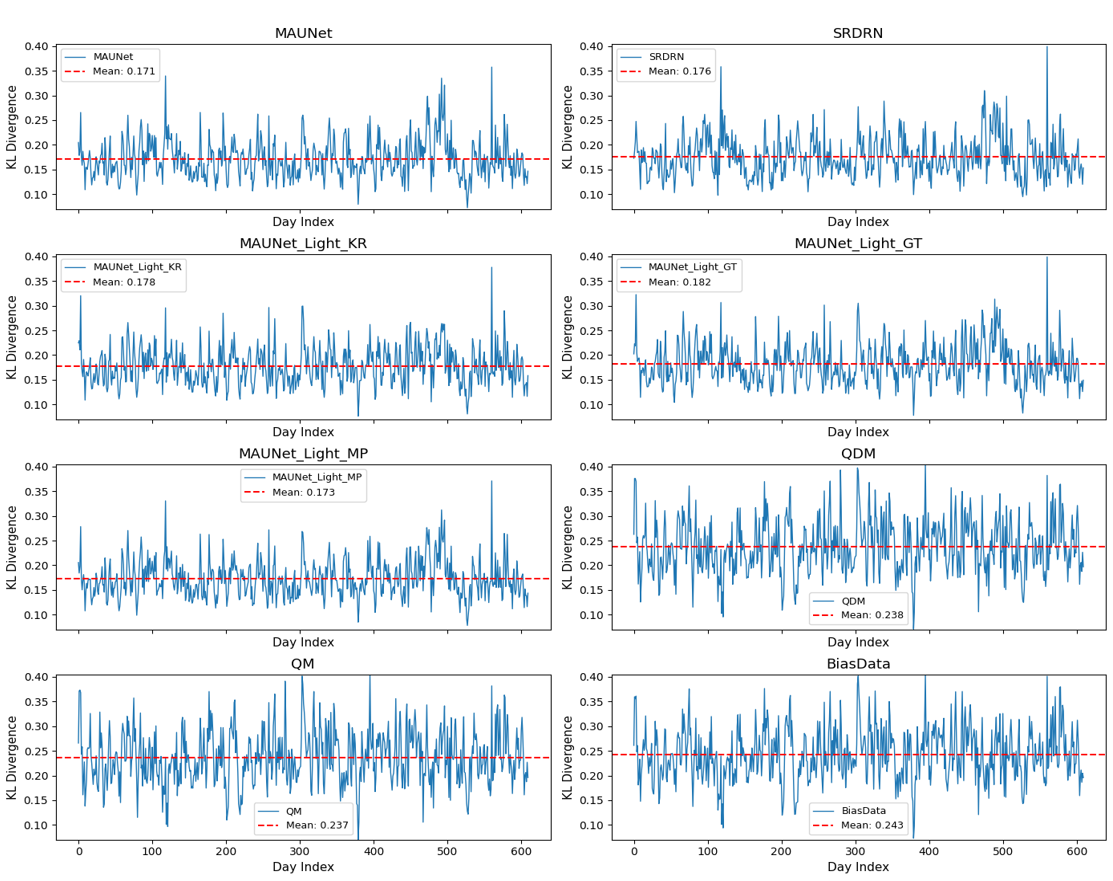}}
\vskip -0.1in
\caption{KL Divergence calculated for daily spatial maps.}
\label{KL_Divergence_Daily_Sample}
\end{center}
\vskip -0.3in
\end{figure*}

After evaluating the model performance in terms of RMSE and correlation coefficient, Kullback–Leibler (KL) divergence was employed to quantitatively assess how well the model reproduces the statistical distribution of observed precipitation. KL divergence measures the discrepancy between the probability distributions of model predictions and observations, thereby providing a distribution-level evaluation. A lower KL divergence indicates that the predicted precipitation distribution closely matches the observed distribution, reflecting the model’s ability to preserve key statistical characteristics such as skewness and variability. KL divergence was computed at each valid grid point by considering precipitation values along the temporal dimension, and the resulting divergence maps are presented in Fig.~\ref{ch7:Gridded_KL_Divergence}. In addition, KL divergence was calculated for daily spatial precipitation fields to examine the divergence of spatial rainfall distributions from the observations, as shown in Fig.~\ref{KL_Divergence_Daily_Sample}. For both analyses, the mean KL divergence values were also computed to facilitate an overall comparison of model performance.

By comparing the KL divergence values across models, this analysis highlights the effectiveness of the proposed approach in capturing the underlying precipitation distribution and reducing distributional bias relative to the observations.

\subsection{Performance Consistency of MAUNet-Light Relative to MAUNet}

\begin{figure}[h!]
\begin{center}
\centerline{\includegraphics[ width=0.9\columnwidth]{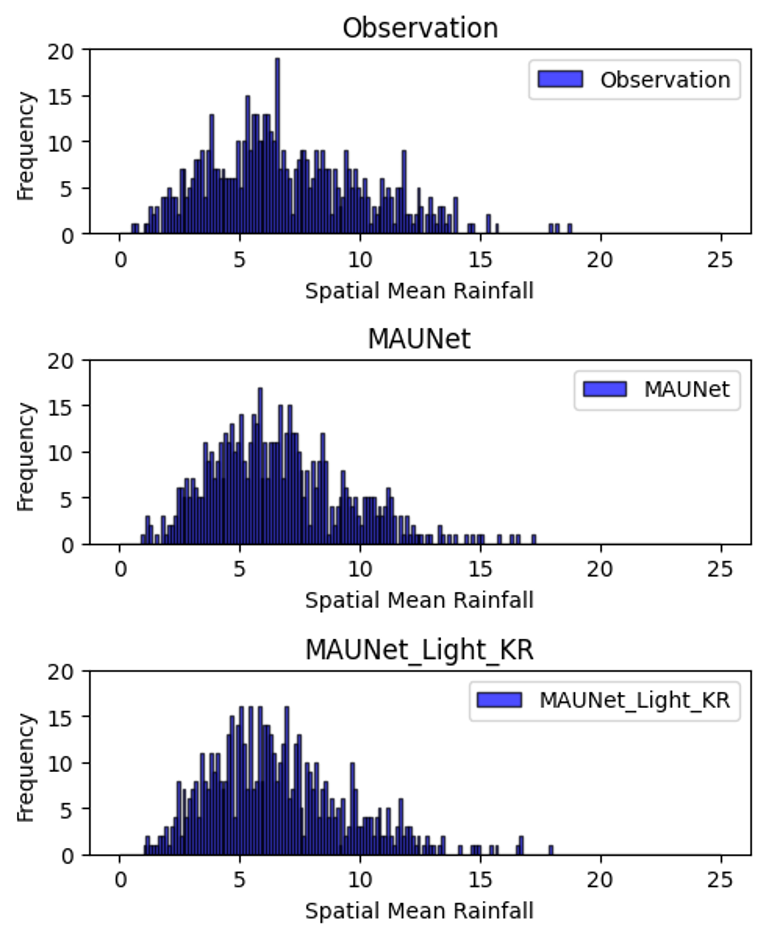}}
\vskip -0.1in
\caption{Comparison of daily spatial mean rainfall distributions predicted by MAUNet and MAUNet-Light with observations.}
\label{SpatialMeanRF}
\end{center}
\vskip -0.3in
\end{figure}

The results of the bias correction and downscaling experiments demonstrate that the MAUNet architecture outperforms other approaches in both tasks. Furthermore, the findings indicate that model compression through knowledge refinement is highly effective in producing lightweight models that substantially reduce computational requirements while retaining performance comparable to that of the teacher MAUNet model. Across all evaluation metrics, MAUNet-Light exhibits performance values that closely match those of the deeper neural architecture, highlighting its ability to achieve an efficient balance between model complexity and predictive accuracy. Fig.~\ref{SpatialMeanRF} illustrates the distribution of spatial mean rainfall values for MAUNet, MAUNet-Light, and the observations. The figure clearly indicates that the precipitation distributions predicted by both MAUNet and MAUNet-Light closely match the observed precipitation distribution.

\subsection{Robustness Analysis of MAUNet-Light}
Sometimes, it is observed that a model trained for a specific task (i.e., producing specific outputs for specific inputs) generates similar outputs even for randomly generated inputs. This indicates model misrepresentation and misleading adaptation, suggesting a lack of robustness. To avoid such misrepresentation and to verify the robustness of the proposed deep learning model, we conducted controlled experiments and analyzed the results. The objective of this experiment is to analyze the model’s behavior when exposed to random noise samples that do not represent real data. Specifically, we examined whether the model produces similar bias-corrected outputs for random inputs as it does for actual input samples.

To carry out this task, random samples were generated using a skew-normal distribution to emulate the inherent statistical characteristics of precipitation. Both temporal and spatial random samples were prepared to analyze model behavior under different sampling strategies. For temporal random sample preparation, each spatial grid cell was processed independently. The mean and standard deviation of the TRMM precipitation time series at each grid cell were computed, and a skewed random time series was generated using the skew-normal distribution. The generated time series was then normalized and rescaled to match the original mean and standard deviation of the TRMM data. For spatial random sample preparation, a similar approach was followed in the spatial domain by iterating over each temporal step (day) in the dataset. For each day, a two-dimensional skewed random field was generated using a skew-normal distribution. The field was subsequently normalized and rescaled to match the observed spatial mean and standard deviation of TRMM data for the corresponding day.

\begin{figure*}[t!]
\begin{center}
\centerline{\includegraphics[width=1.8\columnwidth]{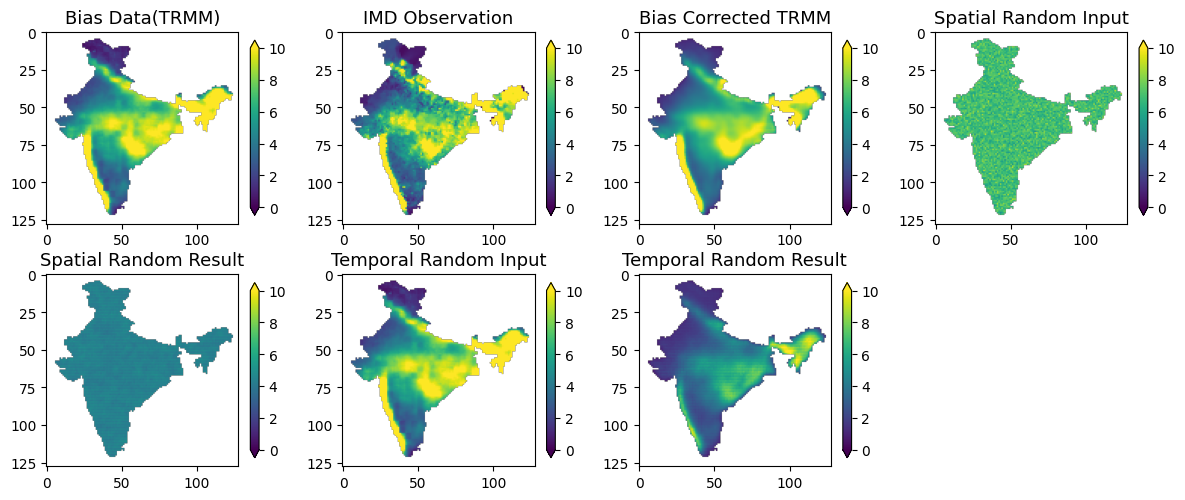}}
\vskip -0.1in
\caption{Gridded mean rainfall maps (Climatology) for the Test period.}
\label{gridded_climatology_random}
\end{center}
\vskip -0.3in
\end{figure*}

\begin{figure*}[b!]
\vskip -0.1in
\begin{center}
\centerline{\includegraphics[ width=1.7\columnwidth]{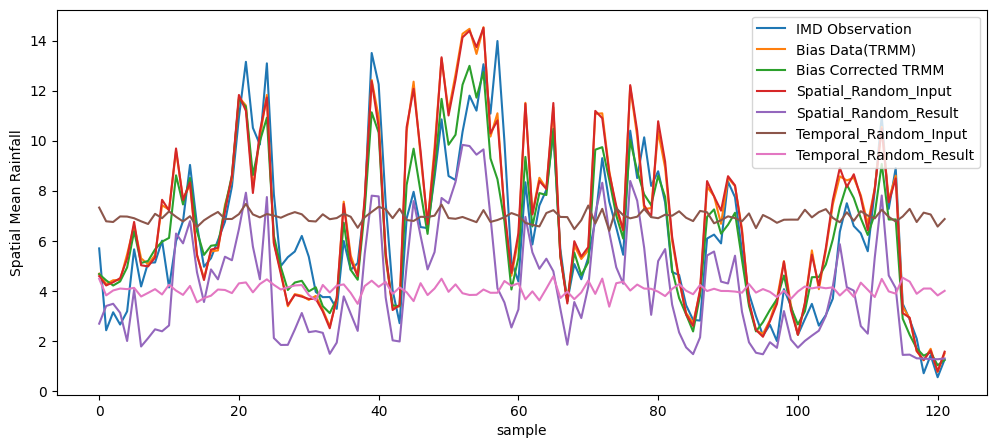}}
\vskip -0.1in
\caption{Daily spatial mean rainfall values for the Year 2015 (ISMR Period).
}
\label{spatial_Mean_RF_Random}
\end{center}
\vskip -0.3in
\end{figure*}

\begin{figure*}[b!]
\vskip -0.1in
\begin{center}
\centerline{\includegraphics[ width=1.7\columnwidth]{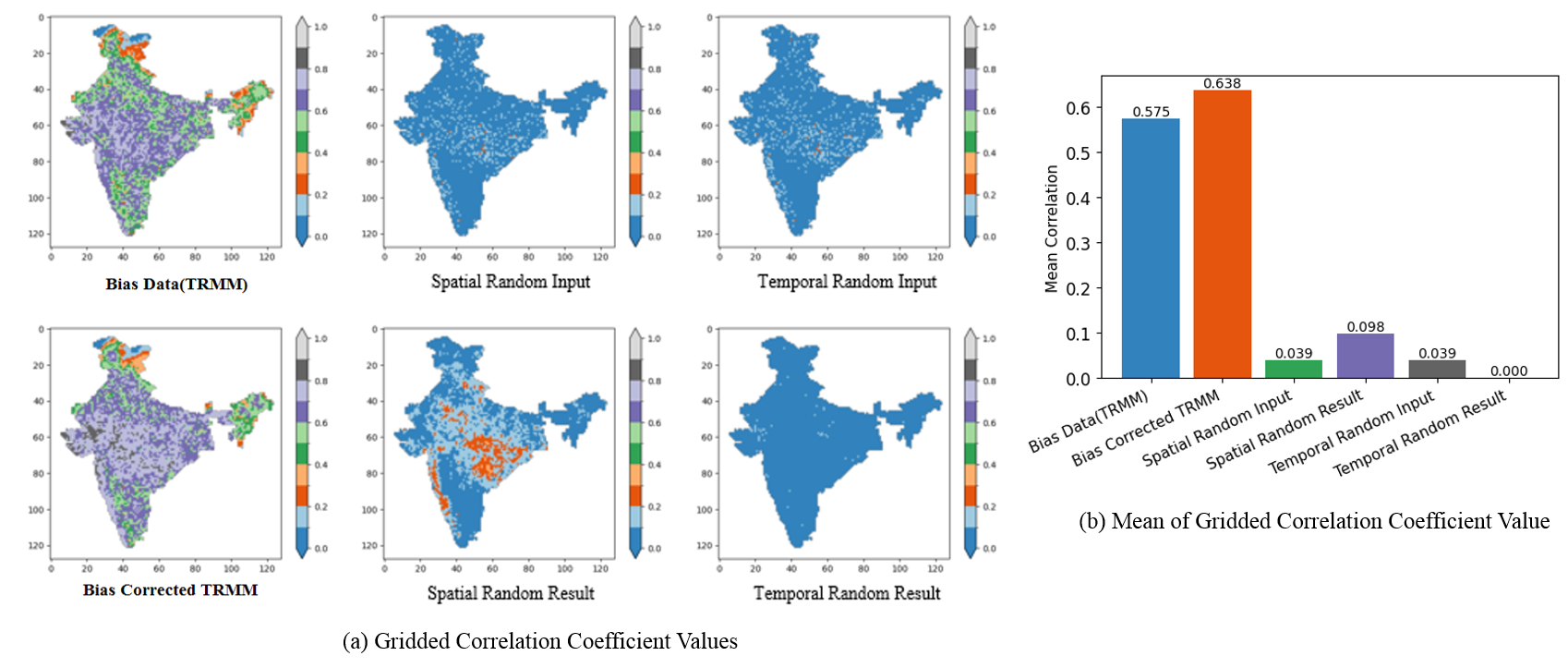}}
\vskip -0.1in
\caption{(a) Plots showing gridded correlation coefficient values calculated between model outputs (or input samples) and observation data, (b) Mean of gridded correlation coefficient values.
}
\label{Correlation_Random_Sample}
\end{center}
\vskip -0.3in
\end{figure*}

After preparing the random samples, they were used as inputs to the trained MAUNet-Light model to obtain the corresponding bias-corrected outputs. These outputs were compared with the observations and bias corrected TRMM data as shown in Fig. \ref{gridded_climatology_random}, \ref{spatial_Mean_RF_Random} and \ref{Correlation_Random_Sample}.

In Fig. \ref{gridded_climatology_random}, the temporal random input samples exhibit a climatological pattern similar to that of the actual bias data, as they were generated by preserving the grid-wise mean and variance of the bias data. Consequently, the temporal random inputs and the bias data naturally share consistent climatological characteristics. Similarly, Fig. \ref{spatial_Mean_RF_Random} shows that the spatial mean rainfall of the spatial random inputs closely resembles the spatial mean of the bias data, since these random samples were generated using the spatial mean and variance of the bias data. 
Figs. \ref{gridded_climatology_random} and \ref{spatial_Mean_RF_Random} clearly demonstrate that the outputs produced from the random samples differ from the actual bias-corrected outputs. Furthermore, the correlation plots and mean correlation values presented in Fig. \ref{Correlation_Random_Sample} clearly indicate that the proposed model is robust, as it does not produce reliable or physically meaningful outputs when driven by random noise inputs. The model successfully differentiates random patterns from meaningful spatial rainfall information, demonstrating its robustness.

\subsection{Representation of Extremes}
\begin{figure}[h!]
\vskip -0.1in
\begin{center}
\centerline{\includegraphics[ width=0.9\columnwidth]{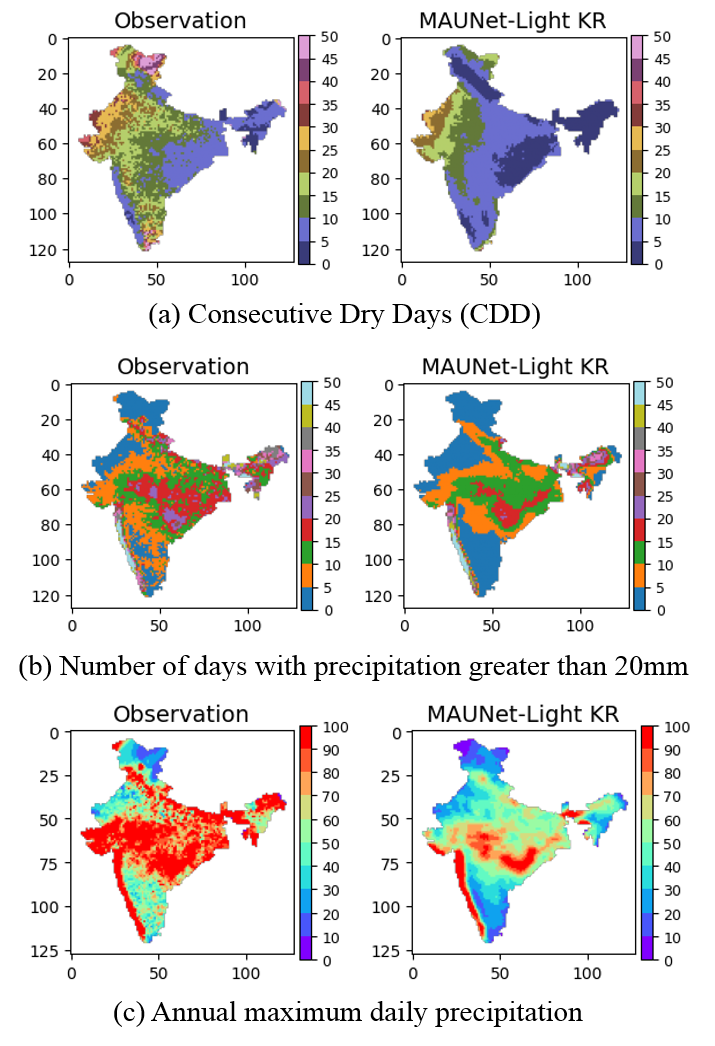}}
\vskip -0.1in
\caption{Extreme Representation: (a) Interannual mean of consecutive dry days(Rainfall $<$ 1mm). (b) Interannual mean of number of days with precipitation greater than 20mm. (c) Interannual mean of annual maximum daily precipitation. }
\label{Extreme_Maps}
\end{center}
\vskip -0.1in
\end{figure}

To effectively represent and evaluate extreme precipitation events, this study employs the precipitation performance framework defined by the Expert Team on Climate Change Detection and Indices (ETCCDI). Specifically, we use indices that quantify consecutive dry days (CDD: Rainfall $<$1mm), the number of heavy precipitation days (precipitation $>$ 20mm), and annual maximum daily precipitation. The spatial distributions of these indices are presented in Fig. \ref{Extreme_Maps}. The results indicate that MAUNet-Light predictions are able to capture the overall patterns of extreme events; however, they still lag behind the observations in terms of magnitude and spatial variability.

\begin{figure}[h!]
\vskip -0.1in
\begin{center}
\centerline{\includegraphics[ width=0.9\columnwidth]{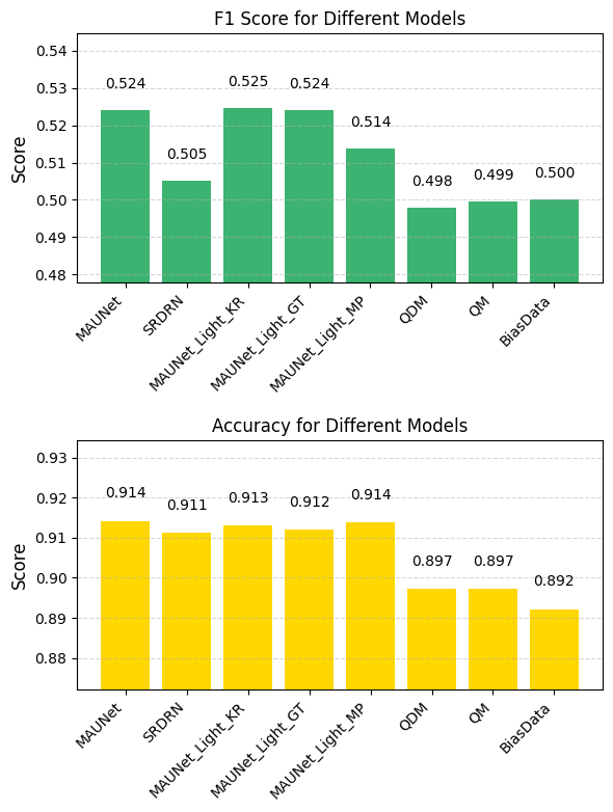}}
\vskip -0.1in
\caption{F1-score and accuracy obtained for different models in capturing extreme rainfall events (RF$\geq$20mm). }
\label{F1_Score_Accuracy_for_Extreme_RF}
\end{center}
\vskip -0.2in
\end{figure}

After examining the extreme precipitation indices, we further assess the ability of different bias-correction models to capture heavy rainfall events. A threshold of 20 mm was applied, where grid-point rainfall values $\geq$ 20 mm were classified as extreme and the remaining values as non-extreme. Based on this binary classification, the F1-score and accuracy were computed for all models, and the results are shown in Fig. \ref{F1_Score_Accuracy_for_Extreme_RF}.

The results indicate that MAUNet-based models consistently outperform the other approaches in detecting extreme rainfall events. Among them, MAUNet-Light-KR achieves the highest F1-score (0.525), demonstrating its superior ability to balance precision and recall. This is particularly important for extreme event detection, where reducing both missed events and false alarms is critical. The accuracy values further support this observation, with MAUNet-Light-KR attaining a high accuracy of 0.913, comparable to the full MAUNet and MAUNet-Light models. In contrast, methods such as SRDRN, QDM, and QM exhibit lower F1-scores and accuracies, indicating weaker performance in capturing extreme rainfall occurrences. Overall, these results highlight MAUNet-Light models as an effective and computationally efficient alternative to the full MAUNet, delivering the best overall performance for extreme rainfall detection.

Although the MAUNet-Light predictions are able to capture the overall spatial patterns of extreme events, they still lag behind the observations. This limitation is consistent with the general tendency of deep neural networks trained under standard settings to converge toward the mean, leading to an under-representation of rare extreme events. These findings highlight the scope for future work aimed at developing more effective lightweight architectures or training strategies specifically designed to improve the representation of precipitation extremes.

\section{Conclusion}
This study demonstrates the effectiveness of deep learning models, particularly MAUNet and its lightweight variant MAUNet\_Light\_KR, in bias correction of satellite precipitation estimates. It also explores the use of MAUNet\_Light in downscaling. MAUNet consistently outperforms other models, while MAUNet\_Light emerges as a good alternate, closely matching MAUNet's performance. The success of the knowledge refinement approach highlights its potential in preparing lightweight models that are computationally efficient and provide a closure performance as that of the deep neural networks. This approach can be utilized to produce multiple lightweight models for multiple tasks. The use of model compression in downscaling and bias correction is relatively rare, and this research can be considered as one of the pioneering works in this field.

\section*{Acknowledgments}
This research was partially supported by the Ministry of Education, Government of India through its SPARC scheme (Sanction order: SPARC/2019-2020/P1585/SL,Dated 26-07-2023). ARG's time was supported in part by the U.S. DOD (SERDP: \#RC20-1183), NASA (21-WATER21-2-0052, Fed. ID: 80NSSC22K1138), and the Indian Monsoon Mission (\#IITM/MM-III/IND-4).


\vspace{-40pt}

\vfill

\end{document}